\documentclass[journal]{IEEEtran}

\usepackage{graphicx}
\usepackage{epsfig}
\usepackage{amsmath}
\usepackage{amssymb}
\usepackage{subfigure}
\usepackage{xcolor}
\usepackage{wrapfig}
\usepackage{chngpage}
\usepackage{lipsum}
\usepackage{tabularx}
\usepackage{array}
\usepackage{adjustbox}
\usepackage{textcomp}
\usepackage{mathtools}
\usepackage[center]{caption}

\DeclarePairedDelimiter\floor{\lfloor}{\rfloor}
\newcolumntype{Y}{>{\raggedleft\arraybackslash}X}

\ifCLASSINFOpdf
\else
\fi

\begin{document}
	
	\title{Binary Document Image Super Resolution for Improved Readability and OCR Performance}
	
	\author{Ram~Krishna~Pandey,~\IEEEmembership{Member~IEEE,}
		K~Vignesh,~A~G~Ramakrishnan,~\IEEEmembership{Senior~Member,~IEEE} and
		Chandrahasa~B
		\IEEEcompsocitemizethanks{\IEEEcompsocthanksitem Ram Krishna Pandey and A G Ramakrishnan are with the Indian Institute Of Science, Bangalore India 560012 E-mail: (ramp@iisc.ac.in and agr@iisc.ac.in) \protect\\
			\IEEEcompsocthanksitem K Vignesh and Chandrahasa B are with IIT Guwahati.
			E-mail: (k.vignesh@iitg.ernet.in and c.bolla@iitg.ernet.in)}
	}
	
	\markboth{}%
	{Shell \MakeLowercase{\textit{et al.}}: Bare Demo of IEEEtran.cls for IEEE Journals}
	
	\maketitle
	
	\begin{abstract}
		There is a need for information retrieval from large collections of low-resolution (LR) \textit{binary} document images, which can be found in digital libraries across the world, where the high-resolution (HR) counterpart is not available. This gives rise to the problem of binary document image super-resolution (BDISR). \textit{The objective of this paper is to address the interesting and challenging problem of super resolution of binary Tamil document images for improved readability and better optical character recognition (OCR)}. We propose multiple deep neural network architectures to address this problem and analyze their performance. The proposed models are all single image super-resolution techniques, which learn a generalized spatial correspondence between the LR and HR binary document images. We employ convolutional layers for feature extraction followed by transposed convolution and sub-pixel convolution layers for upscaling the features. Since the outputs of the neural networks are gray scale, we utilize the advantage of power law transformation as a post-processing technique to improve the character level  pixel connectivity. The performance of our models is evaluated by comparing the OCR accuracies and the mean opinion scores given by human evaluators on LR images and the corresponding model-generated HR images.
	\end{abstract}
	
	
	\IEEEpeerreviewmaketitle
	
	\maketitle
	
	\IEEEdisplaynontitleabstractindextext
	
	\IEEEpeerreviewmaketitle
	
	\section{Introduction}\label{sec:introduction}
	
	\IEEEPARstart{T}{he} task of super resolution (SR) has been a classic problem ever since the earliest work of Tsai~\cite{Tsai}. Digital images are composed of tiny picture elements called ``pixels" and their density in representing the image is commonly referred to as the spatial resolution of that image. Higher the resolution, higher are the details perceivable by the human eyes. The task of SR is said to be ill-posed, since there is no exact and unique solution to the problem and many distinct high-resolution (HR) images can be obtained from a single low-resolution (LR) image. 
	Super resolution techniques come into play, where high resolution is really useful, but the requirement is not satisfied, such as, for example, old fax images. SR techniques enhance the image details and quality by performing non-linear transformations and by eliminating the artifacts generated by the imaging procedures.
	
	Enhancing the resolution of gray images of text has been approached in a variety of ways as presented in \cite{ICDAR} to improve the OCR accuracy. However, the data handled does not encompass the domain of binary document images, which have been scanned at a low-resolution. In the following sections, we have created a unique dataset of Tamil binary document images and propose convolutional neural network (CNN) based SR models for these document images. By exploiting these proposed methodologies, one can improve the readability of the scanned documents even when they are scanned at a low-resolution due to time, hardware or bandwidth constraints. This process eventually leads to better readability and improved accuracy of the optical character recognizer (OCR). These improvements in the perceptual quality and OCR accuracy can be of great value for voluntary initiatives such as ``Project Madurai", which aim to preserve ancient Tamil literature by creating their e-text versions, online \cite{projmdu}. 
	
	It is generally advisable to scan text documents at a resolution of around 300-600 dots per inch (dpi) on a flatbed scanner for the OCR to achieve its best performance. Though it is necessary to scan under these settings, \textit{there already exist large collections of documents which have already been scanned at low resolution and later, the original documents have been destroyed or lost, which prevents us from scanning again}. Also, scanning at a higher resolution implies that we are representing the digitized pixels with more number of dots. This takes time and the files consume a lot of system memory for storage or bandwidth for transmission, thus limiting the user to scan only a few documents in a given period of time and hardware capacity. For example, as mentioned above, in Project Madurai \cite{projmdu}, committed volunteers across the world have shared scanned images of over 500 ancient literary works in Tamil. But, the documents were scanned as binary images at a low-resolution of 75 dpi. This quality was good enough for the volunteers to type in the text by visually inspecting the document image, but is grossly inadequate for the OCR's to give a good recognition performance.
	
	To address the above problem with the objective of achieving higher resolution, mainly two kinds of techniques are used: multiple image super resolution (MISR) and example based super resolution (EBSR). In MISR, we combine multiple images, which are misaligned at the sub-pixel level to obtain a HR image~\cite{MISR1},\cite{MISR2},\cite{MISR3}. Whereas, EBSR refers to improving the resolution of an image by learning the correspondence between low and high resolution image patches from a database~\cite{SISR1}. An improvement over EBSR methodology can be found in \cite{SISR2}, which describes a single image super resolution (SISR) technique for improving the resolution of single LR images. SISR algorithms can be classified based on their field of usage~\cite{benchmark}. Domain-specific SISR algorithms emphasize on a particular type of data: for instance, artwork~\cite{MISR4} and faces~\cite{MISR5}~\cite{MISR6}. Generic SISR algorithms have priors based on common and primitive image properties \cite{SISR3} and can be applied to different kinds of images ~\cite{SISR2}\cite{SISR4}~\cite{SISR8}. Owing to the recent advances in the field of deep learning, we take into account the advantages of the SISR methodology and propose example learning based SISR neural network models, which learn the correspondence between low and high-resolution binary document images and increase the details of any scanned Tamil test document, irrespective of the scanner settings.
	
	Since in super-resolution, our aim is to transform data from the low resolution to high-resolution space, algorithms such as~\cite{VeryDeepCNN,SRCNN} initially upscale the down-sampled image to the required resolution by bicubic interpolation or the combination of interpolations as proposed in~\cite{rkagr} and then pass it through the neural network models. Whereas, models like~\cite{FSRCNN} use transposed convolution at a later stage of the network architecture to upscale the features.
	An alternative to the transposed convolution layer, called the sub-pixel convolution layer \cite{Subpixel}, has also shown promising results for upscaling. Its advantage over the other upscaling layers lies in its computational efficiency and no-training nature. We take into account these factors while building our models and analyze their performance in the latter sections. Here, we list the key-takeaways from our work as follows:
	
	\begin{itemize}
		
		\item The methodologies found in the literature aim to improve the quality of noisy document images in order to achieve higher OCR accuracy. To our knowledge, there is no reported work in the literature on super-resolution of \textit{binary} document images, with the objective criterion of better OCR accuracy and improved readability. There is no standard dataset available for the reported research issue. The ICDAR2015-TextSR dataset~\cite{textdataset}, which is a dataset available for text image super-resolution, contains camera captured gray images. Since there is no publicly available dataset, \textit{we have created our own dataset, which captures  multiple variations in low resolution input images so that the network can generalize well during testing}. The complete details of the created dataset are given in section III.
		
		
		\item 
		We extend the advantages of the synthesis of residual learning and the sub-pixel layer proposed for natural images in \cite{Resnet,Subpixel}, for computationally efficient upscaling of binary document images. Residual learning facilitates the learning of residuals instead of the entire pixel-pixel mapping, which significantly reduces the amount of information that needs to be processed. The usage of sub-pixel convolution layer as the final upscaling function eliminates the need to interpolate the input image in the initial stages, thus allowing the model to learn the complex correspondence between the LR-HR patch pairs.
		
		\item After performing extensive experiments,  we have designed, implemented and proposed 12 different architectures (3 for an upscaling factor of 2 with ReLU as the activation function and 3 more by replacing ReLU with PReLU; similarly 6 more architectures for upscaling the document images by a factor of 4) of various structural complexities, which can reconstruct HR Tamil document images directly from images scanned at a LR. Even though all the architectures perform well in all the cases, our subpixel based architecture outperforms the others.
		
		\item By employing the proposed methodologies, we can store the low-resolution version of a document image (originally scanned at a high resolution) in the system memory and pass it through the model only when we need to perform OCR. Thus, it \textit{saves memory} and facilitates the storage of many more documents. 
		
		\item
		The HR output images of the DNN models have significantly improved \textit{perceptual quality} of the text in the document. This implies that the document now has much better readability than its low-resolution counterpart. The details are given in Section VI. 
		
	\end{itemize}
	
	The following sections are organized as follows: Section II provides information about the related work on super resolution. Section III provides a detailed insight into the dataset creation process. In Section IV, we explain our proposed CNN models, followed by Section V, in which we explain the experimentation process. The results are shown and discussed in section VI, followed by our conclusion in section VII and the possible future work in Section VIII.
	
	\section{Related Work}
	
	The need to improve the details of a low-resolution image for applications in medical imaging, satellite image processing, multimedia, document content retrieval, surveillance, etc. has led to the development of many super-resolution techniques. 
	
	
	Image super-resolution is an ill-posed problem and can be addressed by using priors. The image SR methodologies can be broadly classified into: prediction based~\cite{prediction1}, gradient profile based~\cite{SISR3}, image statistics based~\cite{Statistics1,Statistics2}, patch based models~\cite{Patch1}~\cite{Patch2}, internal~\cite{internal1} and external learning or example based super resolution~\cite{SISR1}. 
	
	Text image super resolution is a domain-specific SISR task, where the training data comprises only the gray document images of the characters of a language. This is unlike the generic SISR methodologies, where the domain is huge. Text image SR methodologies include sparse coding based approaches~\cite{TextSR1,TextSR2,TextSR3,TextSR4}, edge directed tangent fields using Markov random fields~\cite{TextSR1}, Bayesian learning approaches~\cite{TextSR5}, convolutional neural networks ~\cite{TextSR6}~\cite{TextSR7}~\cite{TextSR8}~\cite{rkp,rkagr} and the iterative synthesis of the median estimator~\cite{TextSR9}. 
	
	Internal learning based super-resolution requires the input image itself to reconstruct its high resolution counterpart using the cross-scale self-similarity property~\cite{SISR2}. This property states that small patches of size $ 3 \times 3 $ are highly likely to be found in the down-sampled version of the same image. An example of internal learning based approach is high frequency transfer, wherein the initial HR image is obtained by bicubic interpolation of the input image and the high frequency (HF) components separated from the input image are transferred patch-wise to the bicubic interpolated HR image. Whereas, in the case of neighbor embedding (locally linear embedding), images are super resolved by assuming similarities between the local geometries. It implies that, since there is a correspondence between the LR and HR patches, the HR patches can be obtained by a weighted linear combination of its neighbors by making use of the same weights used with the neighbors of LR patches~\cite{locallinearembedding}. 
	
	On the other hand, example based super-resolution uses a database of LR and HR patch pairs. In this case, the dataset is in a compact format (representation) in terms of the LR and HR dictionaries. Dictionary based approach for SISR was first proposed in~\cite{yang}, and was refined in ~\cite{yimajoint}, which uses sparse coding to find a joint or coupled ~\cite{yimacoupled} representation of LR and HR patches and uses this representation to find the HR patch. Zyde et. al.~\cite{zeyde} use K-SVD for dictionary update and orthogonal matching pursuit for sparse coding. Anchored neighbor regression and its variant ~\cite{Anchoredplus} use smaller dictionaries in place of one larger dictionary to speed up the process. Dong et. al.~\cite{SRCNN} propose CNN based natural image super resolution, which uses bicubic interpolation to resize the images to the same size and then learns a mapping between the resized LR and the corresponding HR images. A lot of improvement in CNN-based super-resolution has been reported in~\cite{VeryDeepCNN}. In ~\cite{DRCNN}, the authors have proposed recursive-supervision and skip connections to make the training easy and to avoid the problems of vanishing and exploding gradients. Another example is the work on fast super resolution CNNs~\cite{FSRCNN}, which has shown restoration quality and speed superior to SR convolutional neural networks~\cite{SRCNN}.
	
	A recent development in image super resolution is photo-realistic SISR using a generative adversarial network (GAN)~\cite{srgan}. They have shown that the super resolved image is of good perceptual quality. However, it is not guaranteed that the network will produce the true high resolution details, as the generator's goal is to fool the discriminator by generating a good quality image similar to the natural image. GAN based approach makes use of a deep CNN, such as VGG-16~\cite{vgg}, for extracting texture information from the images. Thus, when these models are applied to binary document images, which have low texture content, the generator produces symbols, whose structure is retained but not the pixel-connectivity. This goes against the objective of our work.
	
	\textit{We are addressing the problem of SISR for a task distinctly different from the above papers: to enhance the quality of the input binary document images so that the generated images have better readability and OCR accuracy}. Our work builds on what has been reported in~\cite{rkp}~\cite{ram}, which addresses the above mentioned problem for the first time and is able to obtain good PSNR, OCR character and word level accuracies starting from a downsampled version of the document image (gray in nature), the results being similar to that of the corresponding ground truth image. In the current work, we have addressed the \textit{realistic problem of obtaining an upscaled version of a binary document image, actually scanned at a low resolution}. We have achieved this by creating a new dataset and by designing CNNs for binary document image super resolution (BDISR).
	
	\section{Dataset}
	\begin{figure*}[htbp!]
		\centering
		\subfigure[]{%
			\label{hr1}
			\includegraphics[scale = 2.5]{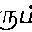}
		}%
		\subfigure[]{%
			\label{lr1}
			\includegraphics[scale = 2.5]{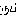}
		}
		\subfigure[]{%
			\label{hr2}
			\includegraphics[scale = 2.5]{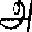}
		}%
		\subfigure[]{%
			\label{lr2}
			\includegraphics[scale = 2.5]{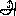}
		}%
		\subfigure[]{%
			\label{hr3}
			\includegraphics[scale = 2.5]{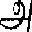}
		}
		\subfigure[]{%
			\label{lr3}
			\includegraphics[scale = 2.5]{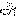}
		}
		\\
		\subfigure[]{%
			\label{hr4}
			\includegraphics[scale = 2.5]{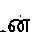}
		}
		\subfigure[]{%
			\label{lr4}
			\includegraphics[scale = 2.5]{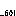}
		}   
		\subfigure[]{%
			\label{hr5}
			\includegraphics[scale = 2.0]{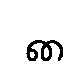}
		}
		\subfigure[]{%
			\label{lr5}
			\includegraphics[scale = 2.0]{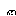}
		}
		
		\caption[]{Magnified sample frames from our dataset. \textbf{(a) }200-dpi high resolution patch \(G_{HR2}\). \textbf{(b)} Low resolution version of (a), generated by taking alternate pixels, \(X_{LR2}\). \textbf{(c)} A 200-dpi high resolution Tamil character, \(G_{char2}\). \textbf{(d)} Low resolution version of (c), \(X_{char2}\). \textbf{(e)} High resolution version of a Tamil character \(G_{HR2-rand}\). \textbf{(f)} Low resolution version of (e), generated by applying the mask of random ones and zeros, \(X_{LR2-rand}\). \textbf{(g)} 150-dpi high resolution image \(G_{150dpi}\). \textbf{(h)} Low resolution image (directly scanned at 75 dpi), corresponding to the image in (g), \(X_{75dpi}\). \textbf{(i)} 300 dpi high resolution image, \(G_{300dpi}\). \textbf{(j)} Low resolution image (directly scanned at 75 dpi), corresponding to the image in (i), \(X_{75dpi}\).
			\label{fig4}}
	\end{figure*}
	
	\subsection{Creation Methodology}
	Our training dataset consists of overlapping patches from the low and high resolution binary images of the same document. We consider them as the training data and the ground truth (GT), respectively. The LR patches are created by taking overlapping patches of stride 1 from the binary LR document image. If an upscaling factor of 'r' is required from the network, we obtain the corresponding HR patches by taking overlapping patches with stride 'r' from the binary HR (GT) image. We have created our dataset under the assumption that a function that upscales by a factor of 2 or 4 is being modeled to super-resolve the LR image. We have created a rich and diverse set of about five million \(LR-HR\) patch pairs, by creating various types of LR images, by choosing alternate pixels, random deletion of pixels, cropping from Tamil character images and direct scanning at low spatial resolution.
	
	\subsubsection{Data for \(2 \times\)upscaling}
	We have scanned Tamil documents at 200 dpi, so that the resulting binary images can be considered as high resolution images for training purposes. Separate copies of these digitized HR document images are now converted to low resolution by selecting only the alternate pixels in the HR image in both x and y directions. Thus, the LR image has one-fourth the pixels of the HR image. We have created around 2 million corresponding pairs of overlapping patches of high and low resolution and train our networks on these patches, instead of the entire image. The dimensions of the LR and HR patches are \(16\times16\) and \(32\times32\), respectively. Since the same document is scanned and converted to a lower resolution, the content of the HR image is reciprocated in the LR image, but with a reduction in the number of pixels, and hence the clarity. Let \(X_{LR2}\) be one of the 2 million LR patches of dimension \(16\times16\) and \(G_{HR2}\) be its HR ground truth of dimension \(32\times32\). The two are related as,
	\begin{align}
	\label{eqn:eqlabel}
	\begin{split}
	X_{LR2}(x,y) = G_{HR2}(2x,2y) ,
	\\
	0\leq x\leq 15, 0\leq y\leq15
	\end{split}
	\end{align} 
	where \(x,y\) are the co-ordinates of the binary \(X_{LR}\) image. Since alternate pixels are considered, the dimensions of \(G_{HR}\) are ensured to be even. Picking alternate pixels to create LR images can be thought of as the scanner skipping the alternate pixels from the original image. As a result, we observe a loss in the structure or shape of the symbols. 
	
	Additionally, we create separate HR and LR patches using a different method. In this procedure, we initially generate the \(LR-HR\) patch pairs following the same methodology as mentioned above, i.e by skipping alternate pixels. Then, we apply a mask, which has randomly distributed ones and zeros on each of the LR patches. Since the distribution is random, the mask for each LR patch is different, but the dimensions of the mask are the same as that of the LR patches, i.e \(16\times16\). The image patch that entails is the result of pixel-wise multiplication of the mask and the LR image patch. Using (1), we get the LR patch through alternate pixel removal. If \(X_{LR2-rand}\) is the patch obtained after applying the mask \(M_{rand}\) on \(X_{LR2}\), then 
	\begin{equation} \label{data2}
	X_{LR2-rand} = (X_{LR2}).(M_{rand})
	\end{equation}
	where the dot operator (.) represents the element-wise multiplication of \(X_{LR2}\) and \(M_{rand}\). \(M_{rand}\) is a matrix of randomly placed ones and zeros. The ones and zeros have been generated by non-uniform probability distributions such as a Gaussian distribution:
	\begin{equation} \label{data2}
	p(x|\mu,\sigma) = \frac{1}{{\sigma \sqrt {2\pi } }}e^{{{ - \left( {x - \mu } \right)^2 } \mathord{\left/ {\vphantom {{ - \left( {x - \mu } \right)^2 } {2\sigma ^2 }}} \right. \kern-\nulldelimiterspace} {2\sigma ^2 }}} 
	\end{equation}
	
	
	The values of $\mu$ and $\sigma$ are initially chosen to be 0, 1 respectively, and are varied to generate different masks resulting in the creation of diverse low-resolution data. One can observe more discontinuities in the pixel structure of \(X_{LR2-rand}\) than in \(X_{LR2}\). This helps the model to be trained in such a way that it can tackle super-resolution tasks of randomly lost pixel data from a document. The ground truth for these new patches is denoted as \(G_{HR2-rand}\).
	
	In order to specifically improve the resolution of the characters (and thus to further enhance the performance of the OCR), we make use of the individual Tamil character data. This data facilitates improvement in pixel connectivity between the strokes in the symbols in the high resolution output. The HR data consists of 200 Tamil symbols, each having 150 samples on an average. Further, each sample is manipulated to create 15 different rotated variants. We follow the two previously mentioned procedures to create the low-resolution patches for this character data also. Let this entire low-dimensional data be denoted as \(X_{char2}\) and the ground truth as \(G_{char2}\).
	
	Finally, to generalize the upscaling function and to make it independent of the input resolution, font and thickness of the symbols in the document image, we create an additional dataset of 2 million LR-HR pairs from images scanned at 75 and 150 dpi, respectively. Let this data be represented as \(X_{75dpi}\) and \(G_{150dpi}\). Our entire dataset for upscaling by two is thus the combination of all the low-resolution data:
	\begin{equation}
	X_{LR} = X_{LR2}+X_{LR2-rand}+X_{char2}+X_{75dpi}
	\end{equation}
	The ground truth data \(G_{HR}\) comprises
	\begin{equation}
	G_{HR} = G_{HR2}+G_{HR2-rand}+G_{char2}+G_{150dpi}
	\end{equation}
	\subsubsection{Data for \(4\times\) upscaling} While generating the data for upscaling by four, the same procedures are used but the 200 dpi images are replaced by 300 dpi images as the ground truth. 
	Our entire dataset for upscaling by four is the combination of low-resolution data comprising:
	\begin{equation}
	X_{LR} = X_{LR4}+X_{LR4-rand}+X_{char4}+X_{75dpi}
	\end{equation}
	and the ground truth comprising:
	\begin{equation}
	G_{HR} = G_{HR4}+G_{HR4-rand}+G_{char4}+G_{300dpi}
	\end{equation}
	
	\subsubsection{Test data}
	OCR performance does not differ much between 300 and 600 dpi images; but its performance on 75 and 100 dpi images is significantly lower than those on higher resolution images. Thus, without loss of generality, we choose the test data to be full length Tamil document images of 75 dpi resolution.
	When we pass these LR images through an OCR, the accuracy with which the OCR predicts the letters is low, resulting in an output of incorrect information. Now, by utilizing this dataset, one can train CNN models to convert LR Tamil images into HR images with better readability and OCR performance. Figure~\ref{fig4} illustrates a few character samples from this dataset, after scaling them for visual clarity.

	\section{PROPOSED CNN MODELS}
	\begin{figure*}[!ht]
		\centering
		\includegraphics[width=0.90\textwidth,height=0.125\textheight]{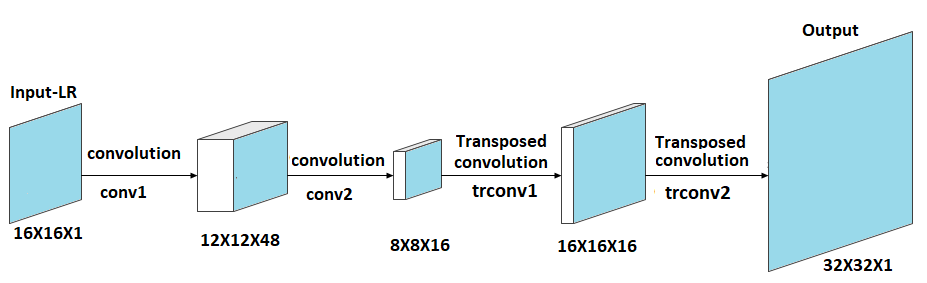}
		\centering
		\caption{Binary document image super resolution using convolution-transposed convolution architecture (CTC)}
		\label{fig: CTC}
	\end{figure*}
	
	We propose six (three, each with ReLU or PReLU as activation functions) neural network models to upscale the binary, low-resolution document images by a factor of two. We have also extended these to obtain six more architectures for upscaling by a factor of four.
	
	\subsection{Convolution-Transposed Convolution Model (CTC)}
	In this convolution-transposed convolution (CTC) architecture, we make use of two convolution layers (conv1 and conv2) without padding, followed by two transposed convolution layers (trconv1 and trconv2) for upscaling the input image \(X_{LR}\) by a factor of two, as shown in Fig.~\ref{fig: CTC}. A deeper model would consume a lot of time to train and test and hence, we have designed this architecture to have a balance between performance and speed. We have also used ReLU and PReLU activation functions and evaluated the performance of the model, as explained in the following sections. In order to upscale by a factor of 4, we add a new transposed convolution layer instead of replicating the entire model again. This approach has been followed in order to reduce the network depth, while achieving an upscale factor of 4.
	
	\subsubsection{Transposed convolution}
	
	A transposed convolution (TC) layer, also called fractionally strided convolution layer, operates by interchanging the forward and backward passes of the convolution process~\cite{Deconvolution}. It has found its application in semantic segmentation~\cite{Segmentation}, representation learning~\cite{Rep-Learning}, mid-level and high-level feature learning~\cite{Features-DC}, etc.
	
	To enhance the resolution, we need a function that maps every pixel in the low-dimensional space to multiple pixels in the high-dimensional space. This can be achieved by introducing the transposed convolution layer after extracting features in the low-dimensional space. Unfortunately, this method has some demerits: the kernel can have uneven overlaps with the input feature map, when the kernel size (the output window size) is not divisible by the stride (spacing between the input neurons)\cite{Distill}. These overlaps occur in two dimensions, resulting in checkerboard-like patterns of varying magnitude. To tackle this issue, we use unit stride, TC layers, along with increasing kernel sizes for our task. There are other alternatives to tackle this problem. For instance, by upscaling the LR image using bilinear interpolation and then utilizing the convolution layers for feature computation, we can prevent the occurrence of these checkerboard patterns. However, naively using this process may lead to the high-frequency image features being ignored during the upscaling \cite{Distill}.
	
	In the proposed architectures, the dimension of the output feature of a transposed convolution layer can be calculated as given in \cite{Deconv-Guide}. According to \cite{Deconv-Guide}, a convolution described by stride s = 1, padding p = 0 and filter size k has an associated TC described by filter size $k' = k$, stride $s' = s$ and padding $p' = k-1$ and its output size \(o'\) is given by,
	\begin{equation} \label{DC}
	o' = i' + (k-1)
	\end{equation}
	where $i'$ is the dimension of the tensor input to the transposed convolution layer.
	
	\subsection{Parallel Stream Convolution Model (PSC)}
	
	Following the previous approach, we add a convolution layer (conv3) to the output of trconv1 layer, whose output is then merged with the input (see Fig.~\ref{fig: PSC}). Since we are merging the feature output with the input image, the dimension of the merged feature map is the same as that of the input. Therefore, we make use of a transposed convolution upscaling layer (trconv3) in the end to upscale by two times. Now, we have two parallel feature maps, which are merged to obtain the final high resolution output as shown in Fig.~\ref{fig: PSC}. So, we call this as parallel stream convolution (PSC) architecture. The performance of this method is also evaluated using ReLU and PReLU activation functions. 
	
	\subsubsection{Residual training}
	Residual learning is most useful, when there is a chance of occurrence of exploding/vanishing gradients while training the network. Simply stacking more layers does not improve the performance of the network, as compared to combining residual blocks of layers. In residual learning, the network does not learn the exact pixel-pixel correspondence; instead, it learns the residual output, which consists mostly of zeros or negligible numbers~\cite{Resnet}. Thus, the network can be trained at a higher learning rate to predict the residuals rather than the actual pixels, while using more number of layers than the usual CNN's ~\cite{VeryDeepCNN,DRCNN,Resnet}. In our \(PSC\) architecture, we have a residual connection from the input to one of the intermediate layers, instead of typically connecting it to the final output layer. Thus, we can represent these connections as,
	\begin{equation} \label{Res}
	f_{res-c3}(X_{LR}) = f_{c3}(X_{LR}) + X_{LR} 
	\end{equation}
	
	where \(f_{res-c3}(X_{LR})\) is the output tensor after merging the output of conv3, \(f_{c3}(X_{LR})\) and the input, \(X_{LR}\). We can now write the outputs of trconv2 and trconv3, in terms of the outputs of trconv1 and conv3 as,
	\begin{equation} \label{Res}
	f_{tc2}(X_{LR}) = W_{tc2}*f_{tc1}(X_{LR}) + b_{tc2} 
	\end{equation}
	\begin{equation} \label{Res}
	f_{tc3}(X_{LR}) = W_{tc3}*f_{res-c3}(X_{LR}) + b_{tc3}  
	\end{equation}
	The predicted output is finally given by,
	\begin{equation} \label{Res}
	f_{f}(X_{LR}) = f_{tc2}(X_{LR}) + f_{tc3}(X_{LR}) 
	\end{equation}

	Since we are combining the input image with the intermediate feature tensor, it is sufficient for the network to learn those extra set of features that are required for efficient upscaling, thus obviating the need to learn the redundant features already present in the input image. Here, we show the effectiveness of using residual connections between the intermediate features instead of initially upscaling the input image and combining it with the CNN model's final output features.
	
	\subsection{CTC-Sub-pixel Convolution Model (CTS)}
	
	In this case, we deploy a ``sub-pixel convolution" upscaling layer to perform the upscaling from the low to the high dimensional feature space. We replace one of the transposed convolution layers (trconv2) with a single sub-pixel layer (SubPixel1) in the same architecture as that of \(CTC\), as shown in Fig.~\ref{fig: CTS}. Since sub-pixel operation does not have trainable parameters as other upscaling layers do, the computational complexity is less than that of \(CTC\). To achieve further upscaling, we need to increase the number of feature maps in the layer before sub-pixel convolution. With this technique, only a single sub-pixel layer is sufficient to upscale by a factor of two or four.
	
	\subsubsection{Sub-pixel convolution}
	
	An alternative to fractionally strided convolution, interpolation and un-pooling methods for increasing the dimensionality is the sub-pixel convolution operation \cite{Subpixel}. This layer is a non-trainable layer, since it only implements matrix manipulations to change the feature dimensions and does not have any weights to learn. Let us assume that we have the input tensor of dimensions \(H \times W \times C.r^2\) to the sub-pixel convolution layer, where \(H\) and \(W\) are the height and width of the tensor, respectively, \(C\) is the number of channels and \(r\) is the upscaling factor that we initially set out to achieve. Now after sub-pixel operation, this feature vector is periodically shuffled to dimensions \(H.r \times W.r \times C\), thus resulting in an upscaled image. Let us first consider the following equation:
	\begin{equation} \label{SP}
	f_{f}(X_{LR}) = f_{sp}(X_{LR}) = PS(f_{tc1}(X_{LR}))
	\end{equation}
	where,
	\begin{equation}
	\centering
	\label{Deconv1}
	f_{tc1}(X_{LR}) = W_{tc1}*f_{c2}(X_{LR}) + b_{tc1}
	\end{equation}
	
	where, \( X_{LR}\) is the input, low resolution, binary document image, \(f_{f}(X_{LR})\) is the final upscaled image of the \(CTS\), \(W_{tc1}\) and \(b_{tc1}\) are the weights and bias of the transposed convolution layer, respectively, and \(f_{c2}(X_{LR})\) is the output feature tensor of the second convolution layer in \(CTS\) model.
	
	Let \(I\) be the input tensor to the sub-pixel convolution layer. Then the periodic shuffling function \(PS\) is given by,
	\begin{equation} 
	\centering
	\label{PS}
	PS(I)_{x,y,c} = I_{\floor{x/r}, \floor{y/r}, C.r.mod(y,r)+C.mod(x,r)+c} 
	\end{equation}  
	where \(x\), \(y\) and \(c\) are the co-ordinates of the periodically shuffled image. For further explanation of the function, please refer to \cite{Subpixel}.
	
	\begin{figure*}[!ht]
		\centering
		\includegraphics[width=0.90\textwidth,height=0.20\textheight]{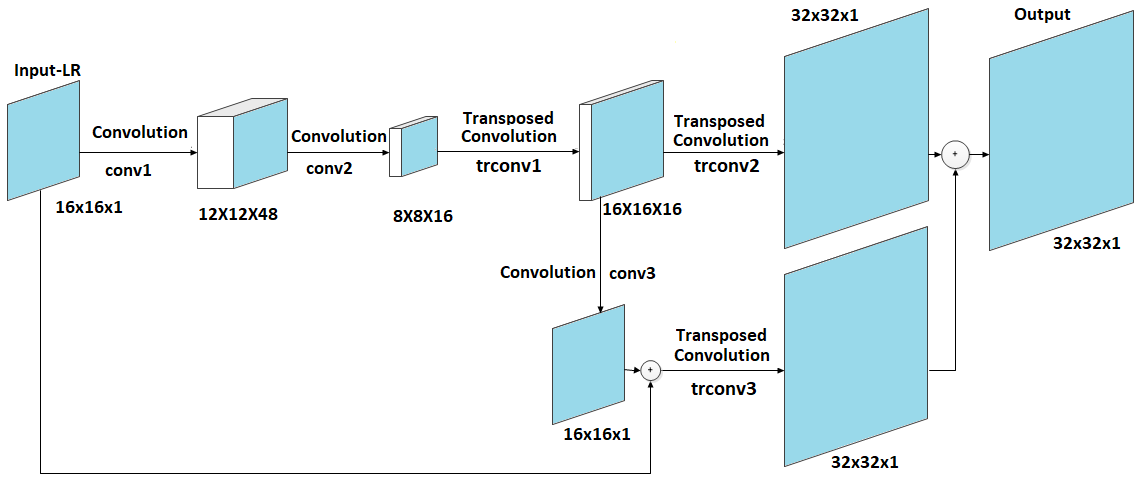}
		\caption{Binary document image super resolution using parallel-stream convolution architecture (PSC)}
		\label{fig: PSC}
	\end{figure*}
	
	\begin{figure*}[!ht]
		\centering
		\includegraphics[width=0.90\textwidth,height=0.125\textheight]{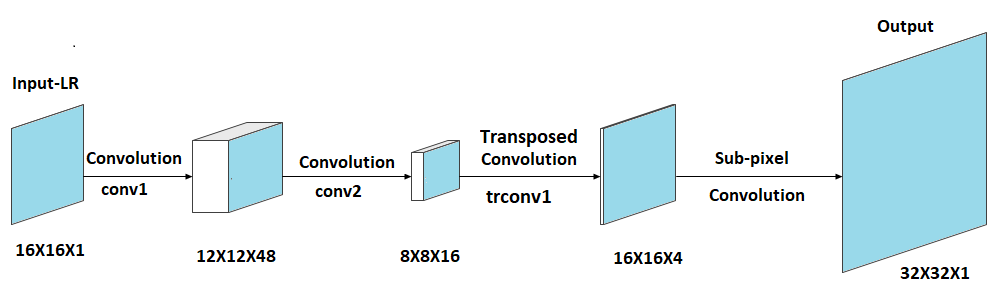}
		\caption{Binary document image super resolution using convolution- transposed convolution-sub-pixel convolution (CTS)}
		\label{fig: CTS}
	\end{figure*}

	\section{EXPERIMENTS}
	
	Our dataset  created consists of approximately 5 million image pairs with diverse low resolution properties for training. The created  image dataset has been saved in compressed \(.npz\) format, with which the training of the different neural network models is performed. 
	
	\subsection{The Activation Functions Used}
	The biologically inspired rectified linear units (ReLU) have been an effective part of neural network architectures since the publication of \cite{ReLU}. ReLU \(f_R(x)=max(0,x)\) converges faster during training than other activation functions and also avoids the vanishing gradient problem. A more generalized (data-dependent) non-linear activation is the PReLU, where the network learns the parameters of the activation function during training \cite{PReLU}. The PReLU function is given by,
	
	\[ 
	f_{PR}(x) =\left \{
	\centering
	\begin{tabular}{ccc}
	$\alpha$\(x\) for \(x < 0\) \\
	\(x \) for \(x >=0\) \\ 
	\end{tabular}
	\right \}
	\]
	
	where $\alpha$ is the data-dependent, learnable parameter.
	
	\subsection{The Loss Function Used}
	
	We use the standard mean square error (MSE) function as the loss function to train the model. 
	\begin{equation} 
	\centering
	\label{MSE}
	Loss = \frac{1}{HWN} \sum_{n=1}^{N}\sum_{x=1}^{H}\sum_{y=1}^{W} (G^{HR}_{n,x,y} - f^{f}_{n,x,y}(X_{LR}))^2
	\end{equation}
	
	Here, \(G^{HR}_{x,y}\) and \(f^{f}_{x,y}(X_{LR})\) refer to the ground truth and the final layer output, respectively, of \(CTC\), \(PSC\) or \(CTS\) models, whichever is being used, at the co-ordinates \((x,y)\).
	\(H,W\) are the height and width of the ground truth/high resolution image, \(N\) is the batch size, and \(n\) is the index of the training data in a particular batch. We optimize the MSE function using the default ADAM optimizer with the following parameter values:  
	$lr=0.001$, $\beta_{1}=0.9$ and $\beta_{2}=0.999.$
	
	\subsection{Implementation Details}
	
	In this subsection, we describe the implementation details of all the models for upscaling by 2 and 4 times.
	\subsubsection{Convolution-transposed convolution architecture}
	The number and sizes of the filters in the CTC model (shown in Fig.~\ref{fig: CTC}) for an upscale factor of 2 are as follows: \(n_{c1} = 48, f_{c1} = 5\times5\), \(n_{c2} = 16, f_{c2} = 5 \times 5\), \(n_{tc1} = 16, f_{tc1} = 9\times9\), \(n_{tc2} = 1, and f_{tc2} = 17\times17\). Table~\ref{table_ctc_2} gives the dimensions of the resulting intermediate feature maps.
	
	Table~\ref{table_ctc_4} gives the dimensions of the feature maps for upscaling by four times. For this, we increase the number of feature maps \(n_{tc2}\) in the second transposed convolution layer from 1 to 8 with filter sizes of \(f_{tc2} = 17\times17\). This is followed by the addition of an extra TC layer (Trconv3) with a depth of \(n_{tc3} = 1\) and filter size of \(f_{tc3} = 33\times33\).
	
	\subsubsection{Parallel stream convolution architecture}
	The number and sizes of the filters in this PSC architecture (shown in Fig.~\ref{fig: PSC}) for upscaling by two are as follows: \(n_{c1} = 48, f_{c1} = 5\times5\), \(n_{c2} = 16, f_{c2} = 5 \times 5\), \(n_{tc1} = 16, f_{tc1} = 9\times9\), \(n_{tc2} = 1, f_{tc2} = 17 \times 17\), \(n_{c3} = 1, f_{c3} = 1 \times 1\), \(n_{tc3} = 1, and f_{tc3} = 17 \times 17\). Table~\ref{table_psc_2} gives the dimensions of the resulting intermediate feature maps.
	
	To upscale by four times, we increase the number of filters in the trconv2 layer: \(n_{tc2} = 8, f_{tc2} = 17\times17\) and add another transposed convolution layer to it, which has \(n_{tc4} = 1\), and \(f_{tc4} = 33\times33\). In the second stream also, we add an extra TC layer, with \(n_{tc5} = 1, f_{tc5} = 33\times33\) and merge the outputs of trconv4 and trconv5 to obtain the final superresolved image. Table~\ref{table_psc_4} gives the dimensions of the feature maps in this case.
	
	\subsubsection{Convolution-transposition-sub-pixel architecture}
	
	The number and sizes of the filters in the CTS architecture (shown in Fig.~\ref{fig: CTS}) for upscaling by two are: \(n_{c1} = 48, f_{cv1} = 5\times5\), \(n_{c2} = 16, f_{c2} = 5 \times 5\), \(n_{tc1} = 4, and f_{tc1} = 9\times9\). This framework is followed by a sub-pixel convolution layer for upscaling by a factor of \(r=2\). Table~\ref{table_cts_2} gives the dimensions of the resulting intermediate feature maps.
	
	To upscale by four times, we increase the number of filters in the TC layer 1: \(n_{tc1} = 48, f_{tc1} = 9\times9\) and add a third convolution layer: \(n_{c3} = 16, f_{c3} = 1\times1\), followed by the sub-pixel convolution layer with upscaling factor \(r=4\). Table~\ref{table_cts_4} gives the dimensions of the different feature maps.
	
	\begin{table}[!ht]
		\centering
		\caption{Dimensions of the intermediate feature maps of CTC model for an upscale factor of 2.}
		\resizebox{0.45\textwidth}{!}
		{
			\begin{tabular}{|c|c|} \hline
				Layer & Dimensions of features (channels last) \\ \hline
				
				Input & 16 $\times$ 16 $\times$ 1  \\ 
				conv1 & 12 $\times$ 12 $\times$ 48  \\ 
				conv2 & 8 $\times$ 8 $\times$ 16  \\ 
				trconv1 & 16 $\times$ 16 $\times$ 16  \\ 
				trconv2 (output) & 32 $\times$ 32 $\times$ 1  \\ \hline
			\end{tabular}
		}
		\label{table_ctc_2}
	\end{table}
	
	\begin{table}[!ht]
		\centering
		\caption{Dimensions of the intermediate feature maps of the CTC architecture for an upscale factor of 4.}
		\resizebox{0.45\textwidth}{!}
		{
			\begin{tabular}{|c|c|} \hline
				Layer & Dimensions of features (channels last) \\ \hline
				
				Input & 16 $\times$ 16 $\times$ 1  \\ 
				conv1 & 12 $\times$ 12 $\times$ 48  \\ 
				conv2 & 8 $\times$ 8 $\times$ 16  \\ 
				trconv1 & 16 $\times$ 16 $\times$ 16  \\ 
				trconv2 & 32 $\times$ 32 $\times$ 8  \\ 
				trconv3(output) & 64 $\times$ 64 $\times$ 1  \\ \hline
			\end{tabular}
		}
		\label{table_ctc_4}
	\end{table}
	
	\begin{table}[!ht]
		\centering
		\caption{Dimensions of the intermediate feature maps of PSC architecture for an upscale factor of 2.}
		\resizebox{0.45\textwidth}{!}
		{
			\begin{tabular}{|c|c|} \hline
				Layer & Dimensions of features (channels last) \\ \hline
				
				Input & 16 $\times$ 16 $\times$ 1  \\ 
				conv1 & 12 $\times$ 12 $\times$ 48  \\ 
				conv2 & 8 $\times$ 8 $\times$ 16  \\ 
				trconv1 & 16 $\times$ 16 $\times$ 16  \\ 
				trconv2 & 32 $\times$ 32 $\times$ 1  \\ 
				conv3 & 16 $\times$ 16 $\times$ 1  \\ 
				trconv3 & 32 $\times $ 32 $ \times $ 1  \\ 
				Output (trconv2+trconv3) & 32 $ \times $ 32 $\times $ 1 \\ \hline
			\end{tabular}
		}
		\label{table_psc_2}
	\end{table}
	
	\begin{table}[!ht]
		\centering
		\caption{Dimensions of the intermediate feature maps of PSC model for upscaling by a factor of 4.}
		\resizebox{0.45\textwidth}{!}
		{
			\begin{tabular}{|c|c|} \hline
				Layer & Dimensions of features (channels last) \\ \hline
				
				Input & 16 $\times$ 16 $\times$ 1  \\ 
				conv1 & 12 $\times$ 12 $\times$ 48  \\ 
				conv2 & 8 $\times$ 8 $\times$ 16  \\ 
				trconv1 & 16 $\times$ 16 $\times$ 16  \\ 
				trconv2 & 32 $\times$ 32 $\times$ 8  \\ 
				trconv4 & 64 $\times$ 64 $\times$ 1  \\ 
				conv3 & 16 $\times$ 16 $\times$ 1  \\ 
				trconv3 & 32 $\times $ 32 $ \times $ 1  \\ 
				trconv5 & 64 $\times$ 64 $\times$ 1  \\ 
				Output (trconv4+trconv5) & 64 $ \times $ 64 $\times $ 1 \\ \hline
			\end{tabular}
		}
		\label{table_psc_4}
	\end{table}
	
	\begin{table}[!ht]
		\centering
		\caption{Dimensions of the intermediate feature maps of the CTS architecture for upscaling by a factor of 2.}
		\resizebox{0.45\textwidth}{!}
		{
			\begin{tabular}{|c|c|} \hline
				Layer & Dimensions of features (channels last) \\ \hline
				
				Input & 16 $\times$ 16 $\times$ 1  \\ 
				conv1 & 12 $\times$ 12 $\times$ 48  \\ 
				conv2 & 8 $\times$ 8 $\times$ 16  \\ 
				trconv1 & 16 $\times $ 16 $\times$ 4  \\ 
				sub-pixel (output) & 32 $\times$ 32 $\times$ 1  \\ \hline
			\end{tabular}
		}
		\label{table_cts_2}
	\end{table}
	
	\begin{table}[!ht]
		\centering
		\caption{Dimensions of the intermediate feature maps of the CTS model for an upscale factor of 4.}
		\resizebox{0.45\textwidth}{!}
		{
			\begin{tabular}{|c|c|} \hline
				Layer & Dimensions of features (channels last) \\ \hline
				
				Input & 16 $\times$ 16 $\times$ 1  \\ 
				conv1 & 12 $\times$ 12 $\times$ 48  \\ 
				conv2 & 8 $\times$ 8 $\times$ 16  \\ 
				trconv1 & 16 $\times $ 16 $\times$ 48  \\ 
				conv3 & 16 $\times $ 16 $\times$ 16  \\ 
				sub-pixel (output) & 64 $\times$ 64 $\times$ 1  \\ \hline
			\end{tabular}
		}
		\label{table_cts_4}
	\end{table}
	
	\subsection{Power-Law Transformation}
	In document images, some characters may split into multiple segments making it unsuitable for the OCR to recognize them properly. Thus, it helps if we can have a method of increasing the spread and connectivity of the pixels in each character before feeding the document image to the OCR. We utilize the power-law transformation to fulfill that need. The basic form of power-law transformation \cite{gammacorrection} is:
	\begin{equation} 
	f_o(x,y) = c f_i^\gamma(x,y)
	\end{equation}
	
	where \(f_i(x,y)\) and \(f_o(x,y)\) are the input and output intensities, respectively, \((x,y)\) are the co-ordinates of the gray-scale images, and \(c\) and $\gamma$ are positive constants. In our case, \(f_i(x,y)\) is the CNN model output. The exponent in the power-law equation is referred to as \(gamma\). Hence, this process was originally called gamma correction. In our
	experiments, $\gamma$ is varied in the range of 0 to 1 in steps of 0.1, while the value
	of c is fixed as 1. When $\gamma$ = 1, the image pixel intensities are unchanged and thus the output undergoes normal binarization, the same as that of the input. When $0<\gamma<1$, we observe that the split characters get merged in the output, which results in better OCR performance. If we increase °$\gamma$ to values higher than 1, the individual split components of characters may further split into multiple components, leading to poorer performance of OCR in recognizing the characters and words in the document images.
	
	\section{Results and Discussion}
	\begin{figure*}[!ht]
		\centering
		\includegraphics[width=0.70\textwidth,height=0.12
		\textheight]{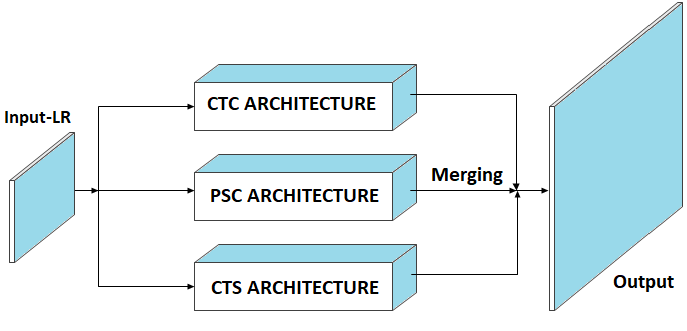}
		\caption{Binary document image super resolution using multi-parallel stream architecture.}
		\label{fig: DISR ARCHITECTURE}
	\end{figure*}
	
	We compare the OCR accuracy on the input low resolution, binary image with those on the images reconstructed by the proposed methods. We consider the OCR accuracy to be the highest priority comparison metric, since it is an objective measure of the quality of any document image. We also obtain the mean opinion score (MOS) from twenty human evaluators, ten each of non-Tamils and native Tamils.
	
\begin{figure*}[!htbp]
		\centering
		\includegraphics[width=.95\textwidth,height=0.88\textheight]{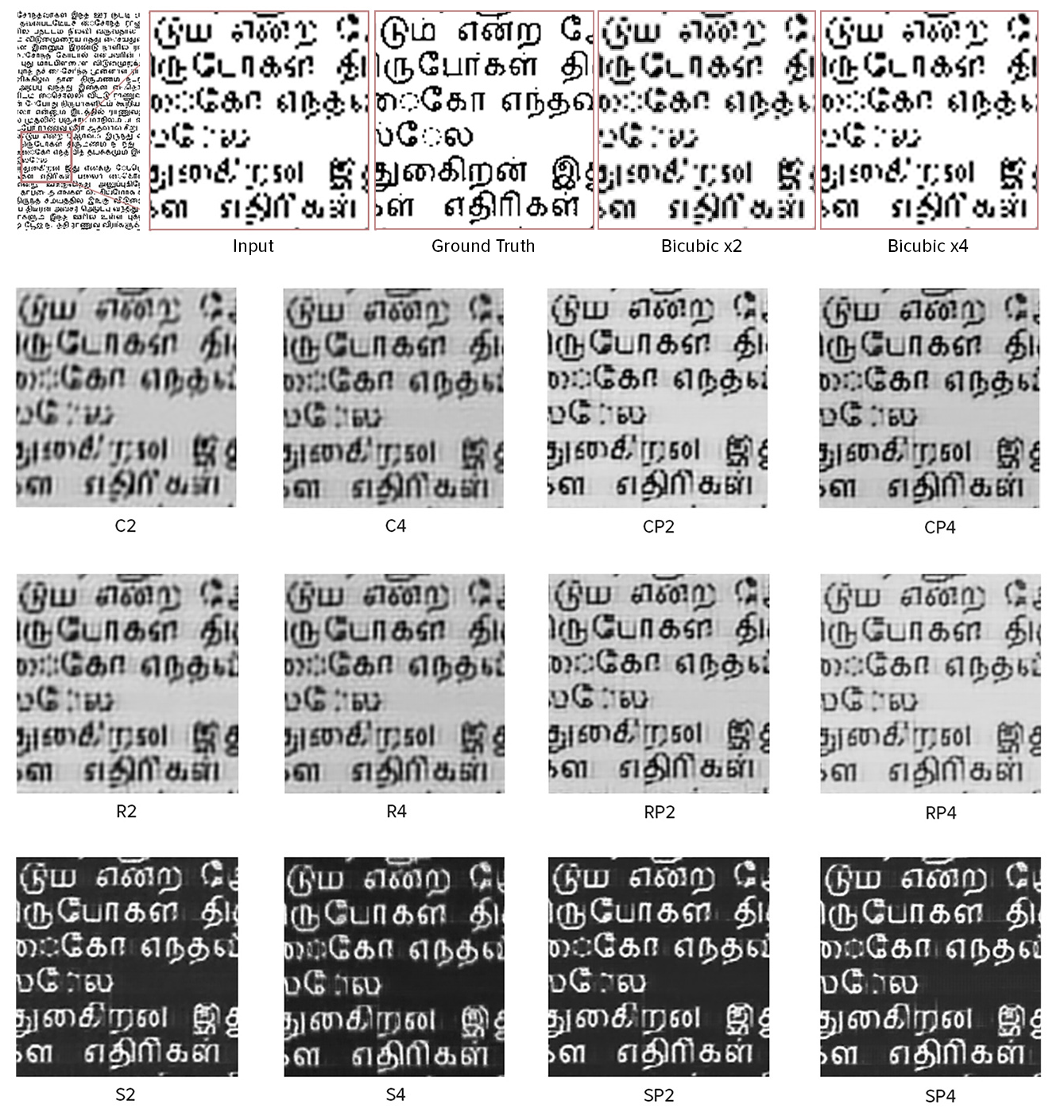}
		\caption{First row depicts the input low resolution, binary image, its corresponding high resolution ground truth and bicubic interpolated images with upscale factors of 2 and 4, respectively. The other rows illustrate the images output by the following models: C2, C4: convolution-transposed convolution architecture, CTC for upscale factors of 2 and 4; CP2, CP4: CTC using PReLU for upscale factors of 2 and 4; R2, R4: Parallel stream convolution architecture for upscale factors of 2 and 4; RP2, RP4: PSC using PReLU for upscale factors of 2 and 4; S2, S4: CT-subpixel convolution architecture for upscale factors of 2 and 4;  SP2, SP4: CTS using PReLU for upscale factors of 2 and 4.}
		\label{fig: output}
	\end{figure*}

	Figure~\ref{fig: output} gives the results of the various proposed models for a small cropped region of one of the input test images. From Fig.~\ref{fig: output}, we can qualitatively observe the major and minor differences in the character level predictions of the proposed models. The left figure in the top panel shows the input, which has been cropped from the Tamil document image and zoomed for the purpose of visualization and beside it is the corresponding zoomed ground truth image. For the sake of visual comparison, we have used bicubic interpolation as a baseline and given the  images interpolated by factors of 2 and 4. The second row displays the outputs of the \(CTC\) model and its variants. The first image C2 is the two times upsacled output of the \(CTC\) model. The second image C4 is the result of four times upscaling. The third result CP2 is the two times upscaled output using PReLU as the activation function. Similarly, CP4 is the image obtained after four times upscaling using PReLU. The third and fourth rows show the output images of the different variants of \(PSC\) and \(CTS\) models, respectively.
	
\begin{figure*}[!htbp]
		\centering
		\includegraphics[width=0.95\textwidth,height=0.88\textheight]{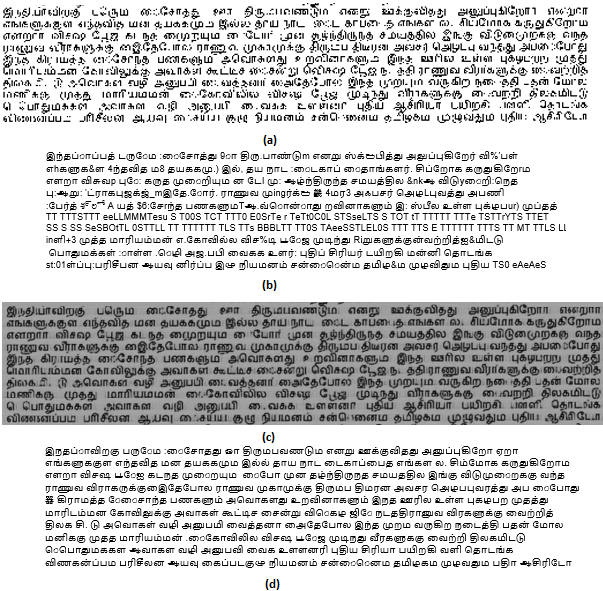}
		\caption{Illustration of the significant improvement in readability and OCR performance of the Tamil binary document image, after enhancement by $4\times$ upscaling using the CTS architecture. (a) A small part of the 75-dpi, binary input image to our CTS model. (b) Output of Google OCR for the input image segment. (c) The corresponding segment of the output image generated by our model. (d) Text output of Google OCR for our generated output image segment.}
		\label{fig: output2}
	\end{figure*}

	Figure~\ref{fig: output2} shows a part of a test image, its output image and the corresponding text outputs obtained from the Google online OCR (Google drive based). Figure~\ref{fig: output2} (a) shows the poor quality of the 75-dpi binary input image, which is not even easy for native Tamils to read directly from. As clearly revealed by the output text given in Fig.~\ref{fig: output2} (b), there are too many errors arising out of the poor image segmentation during the OCR process. Roman and Chinese characters, Indo-Arabic numerals and certain other symbols are wrongly present in the recognized output. Figure~\ref{fig: output2} (c) illustrates the relatively high quality, $4\times$ upscaled image produced by the sub-pixel convolution architecture with PReLU activation. It is obvious that the human readability of the resultant image is high, and that a native Tamil can read the text easily, in spite of some strokes still missing. Accordingly, the text output by Google OCR (shown in Fig.~\ref{fig: output2} (d)) is also significantly better, where not even a single Roman character or numeral is present.

	\begin{table*}[!ht]
		\centering
		\caption{Character level accuracies (\%) obtained by the OCR on the images output by the different proposed techniques.}
		\resizebox{0.85\textwidth}{!}
		{
			\begin{tabular}{|c|c c c c c c|} \hline
				METHOD & Input & ReLU-$\times2$ & PReLU-$\times2$ & ReLU-$\times4$ & PReLU-$\times4$ & $\gamma$-PReLU-$\times4$ \\ \hline
				CTC &25.83  & 28.01 & 34.35& 46.97 & 52.31 &53.46\\ 
				PSC & 25.83 & 31.70& 36.74 &44.46 & 52.19 & 53.1 \\ 
				CTS & 25.83 & 44.09 & 44.27 & 55.08 & 62.06 & 63.68 \\ \hline
			\end{tabular}
		}
		\label{ocraccuracies}
	\end{table*}
	
	\begin{table}[!ht]
		\centering
		\caption{MOS obtained (on a scale of 1 to 10) from 10 human evaluators (non-Tamils) on the images output by the different proposed techniques.}
		\resizebox{0.48\textwidth}{!}{
			\begin{tabular}{|c|c c c c|} \hline
				METHOD & ReLU-$\times2$ & PReLU-$\times2$ & ReLU-$\times4$ & PReLU-$\times4$  \\ \hline
				CTC &  3    &  4.1 & 5.3 & 5.8 \\ 
				PSC &  3.5  &  3.8 & 5.3 & 4.9 \\ 
				CTS &  7.2  &  7.1 & 7.9 & 8.2\\ \hline
			\end{tabular}
		}
		\label{mosnontamil}
	\end{table}
	
	\begin{table}[!ht]
		\centering
		\caption{MOS obtained from 10 human evaluators (native Tamils) on the images output by the different proposed techniques.}
		\resizebox{0.48\textwidth}{!}{
			\begin{tabular}{|c|c c c c|} \hline
				METHOD & ReLU-$\times2$ & PReLU-$\times2$ & ReLU-$\times4$ & PReLU-$\times4$  \\ \hline
				CTC & 4.5 &5.8 & 4.8 &5.8 \\ 
				PSC & 6.1 & 6.3 &5.8 & 5.5 \\ 
				CTS &  8.5 & 9 & 8.3 & 9.6 \\ \hline
			\end{tabular}
		}
		\label{mosnativetamil}
	\end{table}
	
	Table~\ref{ocraccuracies} compares the mean OCR character level accuracies of the outputs. We observe that the performance of sub-pixel stream is on the average nearly 10\% better than the transposed convolution and 8\% better than the resnet connection streams and hence its outputs are illustrated in Fig.~\ref{fig: output} with black background to differentiate them from those of the other methods. CTS-PReLU $2\times$ scaling results in 71.4\% relative improvement in OCR accuracy over the input image, whereas CTC-PReLU achieves 33\%. Results of upscaling by 4 have more image details than those of upscaling by 2, facilitating the recognition of characters for the OCR software and thus achieving higher character level accuracies. A final processing step is the application of power law transformation on the network output. This results in a marginally better recognition due to the improved connectivity of the image pixels.
	$CTC \gamma-PReLU-\times4$ entails a relative improvement in recognition accuracy of 107\%, whereas $CTS\gamma-PReLU-\times4$ achieves a significant relative increase of 146.5\%.
	
	Tables~\ref{mosnontamil} and~\ref{mosnativetamil} list the mean opinion scores for the quality of the output images, given by ten non-Tamils and native Tamils, respectively. Both the groups of people have subjectively rated (the outputs of) CTS to be the best of the three models. Further, barring a few exceptions, the PReLU outputs have been rated to be better than the ReLU outputs. Also, the non-Tamil evaluators, who purely decide based on the image features, have consistently rated the $4 \times$ outputs to be better than the $2 \times$ outputs. 
	
	The primary issue faced by us initially was the unavailability of a diverse dataset that contained corresponding binary patches of low and high resolutions to train the neural networks. Therefore, we have created our own dataset (which will be made publicly available) and built CNN architectures specific to the task in hand. Methods in the literature using convolutional neural networks have been trained and tested on different datasets (derived from natural, colour images or gray level document images) with different input and output dimensions. Thus, an attempt to compare the results of those models on our binary dataset would require modification of their existing architectures, which may fail to demonstrate the maximum potential of the originally proposed models.
	
	While developing different architectures, we have also implemented a three stream, parallel neural network as shown in Fig.~\ref{fig: DISR ARCHITECTURE}, in which the outputs of \(CTC\), \(PSC\) and \(CTS\) are merged to get the final output. We observe that the sub-pixel convolution layer contributes more details than the other two streams to the overall output, while training on either \(2\times\) or \(4\times\) upscaling data.   
	
	In the previous sections, we have mentioned about the poor performance of the OCR on sparsely connected symbols in the document images. The primary reason for this is the following: when a low quality image is passed to the OCR, since the pixels representing a symbol are not properly connected, during the segmentation stage, many symbols are segmented into multiple pieces. Each of these split components is wrongly classified by the OCR as one of the Tamil symbols, leading to the poor classification of the binary document image.
	
	\section{Conclusion}
	
	In this paper, we have proposed effective architectures for the problem of binary document image super-resolution, using artificial neural networks \cite{patent}. We initially build a basic CNN model to perform two times upscaling of the input low resolution image. We progressively modify this architecture by incorporating additional upscaling layers, residually connecting input to the intermediate feature maps, changing the activation function from ReLU to PReLU and by changing the upscaling function from transposed convolution to sub-pixel convolution. We observe that the model employing sub-pixel convolution as the upscaling function and PReLU as the activation function outperforms the other models. Its four time upscaled output image results in a significant relative improvement in OCR accuracy of approximately 140 \%. For further enhancement of the image details, we perform power law transformation on the neural network output and observe a marginal improvement in the OCR accuracy. An additional benefit of the enhanced quality of the image is the \emph{ improved readability of the document content}, thus making it easier for people to read the super-resolved, low quality document image.
	
	\section{Future Work}
	
	We expect to continue this work with a much larger and diverse dataset encompassing various languages to test its scalability to other languages. We will also be working on building more efficient CNN models. 
	
	
	%
	
	
	
	
	\ifCLASSOPTIONcaptionsoff
	\newpage
	\fi
	
	
	
	%
	\pagebreak

\end{document}